**A Reflection on Learning from Data: Epistemology Issues and Limitations**

Ahmad Hammoudeh[a][1], Sara Tedmori[a], Nadim Obeid[b]

a Princess Sumaya University for Technology

b The University of Jordan

---

[1] Corresponding author: E-mail: at.hammoudeh@gmail.com.

# Abstract

Although learning from data is effective and has achieved significant milestones, it has many challenges and limitations. Learning from data starts from observations and then proceeds to broader generalizations. This framework is controversial in science, yet it has achieved remarkable engineering successes. This paper reflects on some epistemological issues and some of the limitations of the knowledge discovered in data. The document discusses the common perception that getting more data is the key to achieving better machine learning models from theoretical and practical perspectives. The paper sheds some light on the shortcomings of using generic mathematical theories to describe the process. It further highlights the need for theories specialized in learning from data. While more data leverages the performance of machine learning models in general, the relation in practice is shown to be logarithmic at its best; After a specific limit, more data stabilize or degrade the machine learning models. Recent work in reinforcement learning showed that the trend is shifting away from data-oriented approaches and relying more on algorithms. The paper concludes that learning from data is hindered by many limitations. Hence an approach that has an intensional orientation is needed.

## Introduction

Since the beginning of the twenty-first century, there has been significant growth in the amount of recorded data accompanied by technological advances that enabled machine learning models to learn from massive amounts of data. This progress has resulted in remarkable successes and business transformations (Manyika et al., 2011). Artificial Intelligence (AI), Big data, and deep learning (DL) have become ubiquitous buzzwords that are commercialized as magical solutions extending these technologies far beyond the current methodological capabilities (Elish & Boyd, 2018). In this context, Yann LeCun, the chief AI scientist at Facebook and the co-recipient of the 2018 Turing Award for his work in DL, likened the robot "Sophia to AI as prestidigitation is to real magic" (Lecun, 2018b). LeCun attacks the AI scam of introducing a puppet that deceives people about AI.

In this paper, learning from data refers to developing ML models that start from seeing data (observations) and then proceed to broader generalizations. This process is also known as "data mining" or "knowledge discovery" (Fayyad, PiatetskyShapiro, & Smyth, 1996). ML is categorized into three learning categories: supervised learning, reinforcement learning (RL), and unsupervised learning. In supervised learning, the agent learns from labeled data (the correct output is known) and aims to develop a model that fits the observed data and can generalize beyond the observed data. In unsupervised learning, the agent does not know the correct answer. Instead, it aims at finding structures in data. On the other side, agents in RL learn by interacting with the environment with the aim of maximizing the received rewards from the environment (Russell & Norvig, 2009).

While collecting data and employing statistics to analyze data have been practiced for centuries (Hald, 1998), the specific techniques of handling big data go back to at least the 1990s. The term "Big data" entered the business through Gartner's report, which defined big data by three Vs: velocity, volume, and variety (Laney, 2009). Gil Press listed ten definitions used in different contexts and then defined big data as "the belief that the more data you have, the more insights and answers will rise automatically from the pool of ones and zeros" (Press, 2014).

This paper aims to (1) present some thoughts on the conceptual framing for learning from data (knowledge discovery in data) (2) investigate the effectiveness of data in leveraging the performance of ML models by considering the relation between the amount of data in two folds: theory and practice, and (3) investigate the effectiveness of data in RL by examining the state-of-the-art engineering directions in RL.

The paper is structured as follows: The conceptual framing for learning from data is discussed in the section titled "The Epistemology of Learning from Data." The need for a theory for learning from data and the limitations of applying generic mathematical theories are highlighted in the section titled "The Need for a Theory to Learn from Data." The practical limits of learning from data are covered in "The Limitations of Data in Practice."

Some challenges associated with relying on generalizations from data are discussed in the section titled "Challenges of Data-Driven Generalisations." The directions of data-oriented approaches in RL are covered in "The Trend in Reinforcement Learning." Finally, conclusions are drawn in "Conclusion."

## The Epistemology of Learning from Data

This section summarizes the epistemology of learning from data and the framework for obtaining knowledge from data.

### Deductive Approach versus Inductive Approach

Learning from data is an inductive approach. The inductive approach is a reasoning principle that starts from observations and then makes broad generalizations (bottom-up). The deductive approach begins with a general statement and then collects observations to test that statement. ML methods start from data representing collected observations and aim to generalize beyond the observed data. Hence learning from data is an inductive approach. The distinction between inductive and deductive approaches is deeply rooted in the history of human knowledge (Copi, Cohen, & Flage, 2016). It has been known since David Hume that induction is unjustifiable (Frické, 2015). Having trillions of examples confirming a statement does not imply the truth of that statement. On the other side, it took time until Godel, in his Incompleteness Theorem, showed that deduction limits do exist (Russell & Norvig, 2009) as there is a true statement that has no proof within the theory. However, the father of nuclear physics, Rutherford, disfavoured the inductive approach, saying, "All science is either physics or stamp collecting." On the same wavelength, the father of modern linguistics, Naom Chomsky, described collecting observations as collecting butterflies, which is not research as the primary concern of research is discovering "explanatory principles" (Norvig, 2017). The philosopher Karl Popper (1902-1994) opposed the inductive approach in the scientific method. Popper described the belief that one can start with pure observations hoping that knowledge will emerge as absurd (Popper, 2014).

Marvin Minsky described the bottom-up approach as a "wrong way" for developing intelligent machines: "I stopped working on neural nets entirely because that was a bottom-up approach, and one could not design a machine to do intelligent things unless one had some sort of a top-down theory of what kind of behavior should that machine have, once you have a clear idea of what it is to be a smart machine, then you can start to figure what kind of machine can support that" (Minsky, 2015).

The history of AI includes many systems developed with a top-down approach. Examples include expert systems, production systems, and general problem solvers. Expert systems are composed of two parts: a) An inference engine deduces new statements by applying the rules to the known statements, and a set of rules and statements (knowledge base). Expert systems emulate the decision-making ability of an expert (human). Although expert systems have many successful applications in diagnosis and control, each expert system is limited to a narrow domain or specific goals. Production systems include a set of rules and a control strategy on how to follow the rules when the states of the world change. The General Problem Solver (GPS) provides a set of processes intended to solve almost any problem using the same algorithm (reasoning mechanism). GPS divides an overall task into subtasks and then solves each task. Applications of Production systems and GPS include theorem proving in geometry or logic, solving puzzles, and robot navigation (Russell & Norvig, 2009). Nonetheless, the bottom-up approach has been successful in tasks at which the top-down approach failed, such as speech recognition, image recognition, music generation, and machine translation (Lecun, Bengio, & Hinton, 2015).

**Rationalism versus Empiricism**

Whether ML models can gain knowledge when learning from big data or not is linked to the philosophical question of how knowledge is gained. In this context, there are two groups; (1) the empiricist who believe that knowledge is derived from experience by generalizations from observations, and (2) the rationalists who believe that knowledge is independent of experience (Peter Markie, 2017). Data-driven approaches are on the side of empiricism.

Anderson (2008) proposed that big data introduces new epistemology and new paradigm shifts where data can drive science and create scientific knowledge (Hey, Tansley, & Tolle, 2009). In the proposed epistemology, correlations are considered enough to extract knowledge, and issues such as the coherence and the interpretability of the model are not needed as before. Instead of the traditional scientific method that starts by developing a hypothesis followed by experiments for testing that hypothesis, the starting point here is the data, and knowledge is discovered from data using algorithms. The fallacies in the proposed inductive data-driven approach are discussed in (Kitchin, 2014, 2016). Rob Kitchen concluded that the empiricist approach has underlying fallacies, but "profitable knowledge" and insightful objectives can be produced within the inductive frame and without the overhead cost associated with searching for explanation and truth.

**Science versus Engineering**

The successes of ML models that learn from big data have been discussed in literature from two perspectives: one is from an engineering perspective, and the other is from the perspective of science. Engineering aims to build things that work well; hence, learning from big data is well recognized and helps in various applications such as business and industry. However, the aim of science is different: the discovery of the underlying knowledge of a phenomenon. Hence, the contribution of big data in science is controversial.

According to Naom Chomsky (Chomsky, 2011), ML models that learn from data have some successes but many failures. Chomsky distinguishes between successful scientific models that provide "insight" and help answer "the deep whys" from models that offer "accurate predictions" but lack the interpretability and give no insight (Norvig, 2017). "The deep whys" include: why things work and why things are the way they are instead of just describing how they are (Wise & Shaffer, 2015))

Norvig (2017) discussed Chomsky's point of view in detail. Although he agreed that engineering successes are not the aim of science, Norvig sees engineering successes as evidence (but not proof) of successful scientific models. While Chomsky is more into scientific knowledge, Norvig values the role of both inductive and deductive approaches in the development of science (Norvig, 2017).

Yann LeCun considers engineering as a branch of science. LeCun distinguishes between "engineering science," which invents new artifacts, and "natural science," which aims at discovering, explaining, and understanding the phenomena. To LeCun, DL is a "physical science" and partially "engineering science." DL is a "physical science" because DL entails understanding the general properties of the artifacts. DL is partial "engineering science" because experimental investigations help in inventing new artifacts. (Lecun, 2018a)

## The Need for a Theory to Learn from Data

Despite the intensive work in figuring out the best practices to build systems that learn from data, little is known about a theory for machine learning. Ali Rahimi likened ML to alchemy due to: a) the lack of a theory and b) ML research is driven by trial and error (Hutson, 2018). This issue has been addressed in a workshop entitled "Deep Learning: Alchemy or Science?" Yann LeCun, in his lecture titled "The epistemology of deep learning," argued that it is not the case that theories always come first and then followed by successful applications. For example, Teletype came before information theory, and the steam engine preceded thermodynamics. LeCun considers the empiricism of the current DL research as a legitimate way of investigation despite its inefficiency. LeCun also insists on the need for a theory of deep learning (Lecun, 2018a).

A workshop named "Theory of Deep Learning" was held as a part of the International Conference on Machine Learning (ICML) conference in 2016 and 2018. Developing a deep learning theory evolved around finding supporting explanations and a deeper understanding of deep learning components (Mianjy, Arora, & Vidal, 2018; Sankar, Srinivasan, & Balasubramanian, 2018). Some researchers tried to link deep learning to theories in other fields, namely Information theory (Vidal, Bruna, Giryes, & Soatto, 2017; Yu, Wickstrøm, Jenssen, & Pr'incipe, 2019), and Mathematics (Achille & Soatto, 2018; Bietti & Mairal, 2017; Vidal et al., 2017).

Basic concepts of ML and data analytics are rooted in mathematics and statistics. Theories from mathematics and statistics guided the development of the models that learn from data (Hastie, Tibshirani, & Friedman, 2001). However, generic mathematical theories do not fully describe the process. Some proposed theories for deep learning have been falsified, such as Information bottleneck theory (Tishby & Zaslavsky, 2015) which extracts information from a random variable relevant to predicting another random variable by identifying an intermediate variable named "bottleneck." Predictions of Information bottleneck theory were found wrong (Kolchinsky, Tracey, & van Kuyk, 2019; Saxe et al., 2018). Hence a theory is essential for learning from data. Researchers seek a theory that can answer questions like which variables to include in a model, what potential confounds, subgroups, or covariates in the data to account for (Wise & Shaffer, 2015), what data to collect, and how to interpret the results (Coveney, Dougherty, & Highfeld, 2016).

The effectiveness of big data can be addressed from a statistics perspective. As the sample size increases, the sample characteristics become closer to the distribution or the entire population. Hence, the learned model can learn, or maybe memorize, the seen examples with a great chance that any test example will be close to the seen ones and as a result, the output can be recalled or estimated correctly. However, this idea relies on memorization more than learning. In contrast, ML aims to create agents that can generalize beyond the training examples (Domingos, 2012). Machine learning models in supervised learning are evaluated by dividing the data into training examples and testing examples. The training set and the testing set are two random variables generated from a single distribution. ML model learns from the training set and is assessed on the testing set. Probability theory tells us that as the number of examples increases, the similarity between the two sets also increases. When the number of examples goes to infinity, the two variables become identical (Ferguson, 2017).

Statisticians rely on the central limit theorem (CLT) to analyze big data (Allende-Alonso, Bouza-Herrera, Rizvi, & Sautto-Vallejo, 2019; Shintani & Umeno, 2018). CLT states that the sum of a number of independent and identically distributed random variables of any kind of distribution approaches the Normal Distribution as the number of variables increases. In the classical form of the theorem, the mean of a random sample $X_1, X_2,..., X_n$ from any distribution of mean $\mu$ and variance $\sigma^2$ converges to the mean of the distribution as the size of the sample n grows. When n goes to infinity, the difference between the sample mean and the population mean approaches zero. CLT implies that statistical and probabilistic methods that work for normal distribution can also work for problems with any other distribution given a large number of samples. Learning from normally distributed variables is easier than handling other distributions (Fischer, 2011). However, Torrecilla and Romo (2018) state that most classical statistical approaches based on CLT and convergence theorems are invalid when applied to big data because the data in practice are far from the standard random variable.

Meng (2018) showed that the difference between the population mean and the sample mean for any dataset is the product of three terms that measure data quality, data quantity, and problem difficulty. As a result, the "bigness" of such Big Data (for population inferences) should be measured by the relative size (the size of the data to the size of the population), not the absolute size of the data alone. "without taking data quality into account, population inferences with Big Data are subject to a Big Data Paradox: the more the data, the surer we fool ourselves" (Meng, 2018)

ML is concerned with developing algorithms that are expected to improve with experience. However, most of these algorithms are based on quantitative reasoning as they rely on training data to achieve their intended objectives. Little effort has been put into qualitative reasoning in this context.

Recently, Defeasible Argumentation (DA) models have proved to be a sound setting to formalize commonsense qualitative and defeasible Reasoning. A Defeasible Logic is a nonmonotonic logic based on the use of logical rules and priorities between them. It is simple, efficient, flexible, and capable of dealing with many nonmonotonic reasoning aspects. Furthermore, a semantic account based on DA models can be provided for defeasible logic (Governatori, Maher, Antoniou, & Billington, 2004). Some studies show that defesible logic is appropriate for reasoning in various applications such as societies of agents, contracts, Semantic Web, and legal reasoning (Governatori, 2005). The idea is motivated by the fact that available knowledge is usually incomplete and uncertain. Defeasible logic (Moubaiddin & Obeid, 2007, 2008, 2009, 2013; Moubaiddin, Salah, & Obeid,

2018; Obeid, 1996, 2000, 2005; Obeid & Moubaiddin, 2009; Obeid & Rao, 2010; Sabri & Obeid, 2016) is appropriate in those situations where we have only partial knowledge of the actual state of affair. Nonmonotonic rule systems offer more expressive capabilities than classical logic rules. Even though numeric attributes present a useful source of information for quantitative reasoning in many applications and knowledge domains, they have been mainly neglected by defeasible reasoning researchers.

The authors believe that there will potentially be some gain if both ML and argumentation frameworks are integrated. We can then have a combination of both analytical and inductive ML methods. The integration should enable tackling some of the drawbacks of each approach separately while keeping their benefits.

**The Limitations of Data in Practice**

When it comes to learning from data, a common perception is that getting more data is the key to achieving better ML models. The question is, to what extent is this apprehension valid? According to Halevy, Norvig, and Pereira (2009), data-driven learning is unreasonably effective in developing models that work well despite some errors and inaccuracy in data. Hence they encouraged collecting more data and letting the data do the job. Later on, one of the authors, Peter Norvig, in a lecture at Stanford (Norvig, 2010), presented the "Data

Threshold," where he explains how the performance enhancement is bounded by a specific limit called data threshold. After the data threshold is reached, no further improvement can be observed. In some cases, degradation in performance was reported when more data was utilized; the reason, according to Norvig, can be attributed to the quality of the added data.

Domingos (2012) classifies the learners according to their capacity to learn from data into two types: those whose representations are of a fixed size, such as linear classifiers, and those whose representations are of variable size. Fixed-size learners do not enhance with more data, but variable-size learners, in theory, can learn any function given a sufficient amount of data. However, this is not the case in practice due to possible learning difficulties such as falling into local optima (Domingos, 2012). Ng (2018) distinguishes neural networks from other variable-size learners. Unlike other ML algorithms such as (decision trees, Support Vector Machine, etc.), deep neural networks can utilize more data by increasing the depth of the network. Hence, ML practitioners used, to quote Pedro Domingos, "a dumb algorithm with more data surpasses an intelligent algorithm with less data" (Domingos, 2012)). Hence, no wonder that DL occupies the state-of-the-art position in various fields (Lecun et al., 2015).

The relation between the development of the performance of ML models with the increase of data is controversial. In computer vision, a degradation in the performance when the size of the dataset was increased to 100 million images was reported by Joulin, van Der Maaten, Jabri, and Vasilache (2016). However, Sun, Shrivastava, Singh, and Gupta (2017) reported an enhancement in the performance when training on a dataset of 300 million images. However, performance enhancement with more data is found to be logarithmic; A model trained on 1 million images achieved an accuracy of 0.76 (Wu, Zhong, & Liu, 2017), while training on 300 million images resulted in an accuracy of 0.79 (Sun et al., 2017).

In Logic, adding more statements can be of no advantage when the additional premises are irrelevant. For example, if Ghassan is interested in learning whether Laila will join an event or not, and he receives information that "Laila will join the event if Salma joins." An additional premise such as "the cat sits on the mat" will not affect Ghassan's knowledge regarding the topic of interest. On the other side, adding premises such as "Salma and Muna will join the event" and "if Muna joins the event, Laila will not join." The additional premises, in this case, hinder Ghassan from reaching a concrete conclusion.

DL successes are attributed to the development of the processing speed (moving from Central Processing Unit (CPU) to Graphics Processing Unit (GPU), the development of the algorithms, and the growth of data. Even though the DL revolution was ignited in 2012 by the availability of an enormous amount of labeled data (ImageNet), getting labeled data remains one of the main challenges of supervised learning. While both the computational power and the algorithms have been in continuous development, the size of the labeled datasets remained stable, according to Sun et al. (2017).

Overall, deep neural networks keep performing better with more data. However, relying on data to leverage the performance of ML models is hindered by three main challenges:

- the difficulty of getting an enormous amount of labeled data. The recent successes of deep learning are supervised, which entails a need for labeled data. However, the availability of labeled data is a difficult challenge.

- Avoiding the data threshold and reaping the benefits of more data entails optimizing algorithms and data.

- The logarithmic relation between data and performance.

**Challenges of Data-Driven Generalisations**

Obtaining knowledge from data is found to be complicated even for scientists and domain experts. Hughes (2013) discussed an example that demonstrates a long dispute (Flegal, Kit, Orpana, & Graubard, 2013; Willett, Hu, & Thun, 2013) between scientists due to contradictory conclusions obtained from data by domain experts. The reported case concerns the relation between weight and life expectancy; simply getting insights from data led to conflicts in determining if being fat is associated with lower life expectancy or higher life expectancy! The odds of reaching a false conclusion from data are high even when experts draw data-driven conclusions. This can be attributed to confounding factors, data collection biases, and biases in data selection (Hughes, 2013). Hence, obtaining accurate knowledge from data is even more difficult for a machine than for a human expert.

Developing knowledge/generalizations from data has many challenges that include:

- Not all data are reliable; datasets may contain predominantly bad data hence evidence of good experimental design is necessary (Coveney et al., 2016). For example, having records of identical features that yielded different outcomes hinders the possibility of drawing a generalization from data.

- Extrapolating beyond the range of observed data (Coveney et al., 2016): this is demonstrated by the failure of Google's Flu project, which predicted in real-time the advent of a flu epidemic from search queries. However, the project resulted in false predictions when it was deployed and was terminated later. Although Google flu achieved high accuracy on testing data, it failed to extrapolate beyond that data (Butler, 2013).

- Spurious correlations: correlation does not imply dependency (Coveney et al., 2016), and random correlations are possible. Many examples of spurious correlations are available at Vigen (2015). An example of an ML model that won the KDD competition (Knowledge Discovery and Data Mining competition organized by ACM) for breast cancer detection using completely irrelevant data is reported (Perlich et al., 2008). The target class was highly correlated with the patient ID in both the training and testing data.

- Sensitivity to chaos and the tiny errors in data: Szegedy, Christian Zaremba, Wojciech Sutskever, Bruna, Erhan, Dumitru Goodfellow, and Fergus (2014) discovered that many ML models are vulnerable to adversarial examples. A slight change to the input can lead ML models to misclassify a correctly assigned example. For instance, fooling deep neural networks by modifying one pixel was reported in Su, Vargas, and Sakurai (2019). More examples can be found in Akhtar and Mian (2018); Goodfellow, Shlens, and Szegedy (2015).

- Interpretability (black box problem): Although ML models that learn from big data can provide accurate results and excel in their tasks, the reasoning behind the model's outcome remains unclear. Scientists still cannot fully understand or trace the logic behind assigning a particular outcome to a specific input. The lack of interpretability may not be an issue for an ML model that plays chess, but it is a critical issue in medical decisions that affect patients' lives.

Although Holm (2019) agrees that the black box is a problem for engineering and science, Elizabeth encourages accepting the black box, especially when the advantage of utilizing the black box far exceeds the alternatives. Geoffrey Hinton (Simonite, 2018) argued that trusting DL has no difference from trusting humans, as we trust humans even though they are black-boxes.

Away from the debate between academics and philosophers, the demand for "meaningful explanations of the logic involved" turned out to be the law after the adoption of the General Data Protection Regulation (GDPR) by the European Parliament in 2018 (Dumas, 2019). The methods proposed to overcome the interpretability problem have been surveyed recently in Guidotti et al. (2018).

**The Trend in Reinforcement Learning**

In RL, the agent learns from experience by interacting with its environment. The agent learns the strategy that maximizes its reward by taking different actions and facing successes and failures. In contrast with supervised learning, the agent is not informed which actions should be taken. Instead, the agent discovers the action that needs to be taken at each state. For more about RL, the reader is referred to *Sutton2018* (n.d.); Hammoudeh (2018). An RL agent aims to find a strategy that maximizes its rewards. The strategy is developed by interacting with the environment, taking actions, and evaluating the rewards; Actions that led to rewards are promoted, and actions that led to penalties are discouraged. The strategy can be considered, in its simplest form, a set of rules (if the current state is s1, then take action a1). However, for a simple 8x8 board game like Chess, the conservative lower bound number of variations is estimated by Shannon to be $10^{120}$. As this number seems large, the raised question is whether tabular rules can tackle this problem or not. The answer is no. By the crude assumption that the state of each game can be represented by a single bit and saving that bit requires a single atom, the number of atoms needed for chess will be $10^{120}$, which is more than the number of atoms in the observable universe $10^{80}$. Hence approximations and more intelligent solutions are required.

The problem of the enormous amount of data is overcome by replacing the tabular form of a function that takes the state, for example, as input and then estimates the value of interest as an output. A simple function, let's say (y = 2x), embeds an infinite amount of data in few symbols. In other words, one can say that enormous tables can be compressed into functions. As the relations between inputs and outputs get complicated, more complex functions are needed. The behavior of the agent depends on the strategy that the agent develops through the learning process. Hence the learning process can be seen as a function approximation process (*Sutton2018*, n.d.). h Representing the strategy using a function dramatically reduces the complexity of the learning process; For example, a linear function (y =ax+b) can be estimated using two data points only. However, complex strategies entail complex functions with a great number of parameters. In this place, deep neural networks are considered universal approximators that can approximate any function. The idea of using DL as a function approximator for RL strategy has achieved a breakthrough in RL and allowed the development of an agent that is comparable to human-level performance in different tasks. The findings were published in Mnih et al. (2015).

The role of data in RL is not as pivotal as it is in supervised learning. However, the development of supervised learning has led to remarkable successes in RL. In the following paragraphs, the trends of the state-of-the-art RL published in top-ranked journals (H-index greater than 1000, e.g., Nature and Science) were surveyed.

The state-of-the-art RL agent that mastered the game of GO in 2016, Alpha Go (Silver et al., 2016), was trained initially using a database of 30 million moves of human experts, then it developed its skills by playing against itself. However, RL agents, since 2017, have been shifting towards:

- Instead of relying on labeled data that teach an agent at the beginning of the learning process how a professional human would play, the shift is towards an agent that starts from scratch by playing against itself with no supervised learning and no learning from data. Although this is a bottom-up approach as learning occurs by trial and error. However, agents that learn entirely by trial and error were shown to surpass agents that initially start learning from data (Silver et al., 2017).

- Simplicity and less computational power by utilizing algorithmic advances instead of adding more GPUs and TPUs: For example, using a new RL algorithm that incorporates lookahead search inside the training loop resulted in rapid improvement and precise and stable learning (Silver et al., 2018).

AlphZero (Silver et al., 2018) runs on 4 TPUs only to search 10,000's of moves per decision which is less than the best Chess engine (10,000,000's of moves per decision). However, this development remains far from a human grandmaster who plans 100's of moves per decision only. After succeeding in developing a program that outperforms any other program or human without going for more computations, David Silver said: "People tend to assume that ML is all about big data and a massive amount of computation, but what we saw in AlphaGo zero is that algorithms matter much more..." (Silver, 2017)

**Conclusion**

Learning from big data is powerful in achieving engineering success, but it has many epistemological problems and limitations. In its epistemology, learning from data follows an inductive approach. It starts from observations and then draws generalizations; generalizations that fit large data are useful for business and industry, but their contribution to science and research is controversial.

ML is concerned with developing algorithms that are expected to improve with experience. However, most of these algorithms are based on quantitative reasoning as they rely on training data to achieve their intended objectives. Little effort has been put into qualitative reasoning in this context.

Recently, defeasible reasoning (defeasible argumentation) has proved to be a sound setting to formalize commonsense qualitative reasoning. Even though numeric attributes present a useful source of information for quantitative reasoning in many applications and knowledge domains, they have been mostly neglected by defeasible reasoning researchers. A theory for learning from data is essential. Although the mathematical and statistical theories influence the development of ML models and big data analytics, the generic mathematical theories do not describe learning from data precisely.

The increase of data does not always enhance the performance of ML models. Cases for degradation in the performance were reported (Domingos, 2012; Joulin et al., 2016; Norvig, 2010). In practice, the performance development with the increase of the data is shown to be logarithmic at its best. As we get more data, the development of the performance gets slower, and after a specific limit, it stabilizes or maybe starts degrading. Moreover, generalizations obtained from big data suffer from problems such as spurious correlations, the lack of interpretability, and the incompatibility with laws. However, there is good progress to overcome these challenges.

Examining the state-of-the-art in RL showed that the trend is shifting away from data-oriented approaches and moving closer towards intelligent algorithmic approaches. Overall, learning from data has many engineering successes, but it suffers from epistemological challenges and has many limitations as a bottom-up approach. On the other hand, a top-down approach is more efficient and widely established in science. However, its performance is not comparable to recent achievements of the bottom-up approach (namely, deep learning). Hence, a hybrid approach that combines top-down and bottom-up approaches can help reinforce the weaknesses and sharpen the strengths.